\documentclass{article}
\usepackage[T1]{fontenc}
\usepackage{iftex}
\ifPDFTeX\usepackage[utf8]{inputenc}\fi
\usepackage{colm2024_conference}
\usepackage[english]{babel}
\usepackage{amsmath,amssymb}
\usepackage{graphicx}
\usepackage{booktabs,multirow,array,tabularx}
\usepackage{colortbl}
\usepackage{enumitem}
\usepackage{needspace}
\usepackage{wrapfig}
\usepackage{placeins}
\usepackage{fvextra}
\usepackage{tcolorbox}
\tcbuselibrary{breakable,skins}
\usepackage{caption,subcaption}
\usepackage{xurl}
\usepackage{hyperref}
\definecolor{darkblue}{RGB}{0,45,95}
\hypersetup{colorlinks=true,citecolor=darkblue,linkcolor=darkblue,urlcolor=darkblue}
\usepackage[capitalize,noabbrev]{cleveref}
\usepackage{tikz}
\usetikzlibrary{positioning,arrows.meta,calc,fit,backgrounds}
\definecolor{peink}{HTML}{203140}
\definecolor{pegray}{HTML}{687785}
\definecolor{peline}{HTML}{DCE3E9}
\definecolor{peblue}{HTML}{606CB4}
\definecolor{pegreen}{HTML}{31877F}
\definecolor{peorange}{HTML}{BC7A3C}
\definecolor{pepaper}{HTML}{F7F9FC}
\definecolor{penight}{HTML}{183344}
\definecolor{pegold}{HTML}{F1C57C}
\tikzset{
 pebase/.style={font=\sffamily\fontsize{8}{10}\selectfont,text=peink,line cap=round,line join=round},
 peflow/.style={-{Stealth[length=1.8mm,width=1.3mm]},draw=pegray!85,line width=.65pt},
 pefeedback/.style={peflow,draw=peblue,dashed},
 peheading/.style={font=\sffamily\bfseries\fontsize{10}{12}\selectfont,anchor=west,text=peink},
 pelabel/.style={font=\sffamily\bfseries\fontsize{8}{10}\selectfont,text=peink,align=center},
 penote/.style={font=\sffamily\fontsize{7.5}{9}\selectfont,text=pegray,align=center},
 pesmall/.style={font=\sffamily\fontsize{7}{8.5}\selectfont,text=pegray,align=center},
 pecard/.style={rounded corners=2pt,draw=peline,fill=white,line width=.6pt},
 pebox/.style={pecard,align=center,inner sep=5pt},
 pedata/.style={pebox},pemodel/.style={pebox,draw=peblue!65,fill=peblue!5},
 pevisual/.style={pebox,draw=pegreen!60,fill=pegreen!4},
 pejudge/.style={pebox,draw=peorange!60,fill=peorange!4},
 petitle/.style={peheading}
}

\newcommand{\PEPoster}[2]{%
\begin{scope}[scale=#1,transform shape]
 \colorlet{psky}{penight}\colorlet{pglow}{pegold}\colorlet{pstall}{peorange}
 \ifnum#2=2\colorlet{psky}{pegreen!65!peink}\colorlet{pglow}{white!55!pegold}\fi
 \ifnum#2=3\colorlet{psky}{peblue!65!peink}\colorlet{pstall}{peorange!60!red}\fi
 \ifnum#2=4\colorlet{psky}{peink!90!black}\colorlet{pglow}{peorange!50!white}\fi
 \begin{scope}
 \clip[rounded corners=.035cm] (0,0) rectangle (1,1.4);
 \fill[psky] (0,0) rectangle (1,1.4);
 \fill[white,opacity=.035] (0,.65) -- (1,1.35) -- (1,0) -- (0,0) -- cycle;
 \fill[pglow,opacity=.055] (.7,.57) circle (.38);
 \fill[white,opacity=.065] (.03,.30) rectangle (.18,.89);
 \fill[white,opacity=.04] (.19,.30) rectangle (.34,.79);
 \fill[white,opacity=.055] (.79,.32) rectangle (.96,.95);
 \fill[pglow] (.855,1.205) circle (.065);
 \fill[psky] (.88,1.226) circle (.06);
 \node[anchor=west,inner sep=0,text=pglow,font=\sffamily\bfseries\fontsize{4.1}{4.1}\selectfont] at (.10,1.205) {NIGHT};
 \node[anchor=west,inner sep=0,text=pglow,font=\sffamily\bfseries\fontsize{4.1}{4.1}\selectfont] at (.10,1.065) {MARKET};
 \draw[pglow!65!psky,line width=.18pt] (-.02,.935) .. controls (.30,.785) and (.72,.795) .. (1.02,.95);
 \foreach \x/\y in {.13/.875,.31/.837,.50/.822,.69/.84,.88/.889}{
  \draw[pglow!50!psky,line width=.15pt] (\x,\y)--++(0,-.045);
  \fill[pglow,opacity=.10] (\x,\y-.07) circle (.06);
  \fill[pglow] (\x,\y-.07) circle (.018);
 }
 \fill[psky!78!black] (.38,0)--(.64,0)--(.56,.54)--(.50,.54)--cycle;
 \fill[pglow,opacity=.10] (.31,0)--(.82,0)--(.58,.47)--(.49,.47)--cycle;
 \fill[pstall!40!psky] (.055,.13) rectangle (.43,.49);
 \fill[pglow,opacity=.73] (.08,.25) rectangle (.405,.445);
 \fill[psky!65!black] (.11,.25) rectangle (.12,.445);
 \fill[psky!65!black] (.30,.25) rectangle (.31,.445);
 \fill[pstall] (.045,.49)--(.105,.60)--(.37,.60)--(.44,.49)--cycle;
 \foreach \x in {.09,.21,.33}{\fill[pglow,opacity=.68] (\x,.49)--(\x+.035,.60)--(\x+.075,.60)--(\x+.065,.49)--cycle;}
 \fill[pglow!65!pstall] (.06,.205) rectangle (.43,.26);
 \fill[pstall!50!psky] (.635,.22) rectangle (.965,.59);
 \fill[pglow,opacity=.65] (.66,.34) rectangle (.94,.57);
 \fill[pstall!85!pglow] (.61,.59)--(.67,.70)--(.92,.70)--(.995,.59)--cycle;
 \foreach \x in {.66,.78,.9}{\fill[pglow,opacity=.72] (\x,.59)--(\x+.024,.70)--(\x+.06,.70)--(\x+.056,.59)--cycle;}
 \fill[pglow!80!pstall] (.635,.30) rectangle (.965,.35);
 \foreach \x/\y/\s in {.46/.135/1,.59/.30/.76,.80/.10/1.15}{
  \begin{scope}[shift={(\x,\y)},scale=\s]
   \fill[psky!75!black] (0,.09) circle (.024);
   \fill[psky!75!black,rounded corners=.012cm] (-.027,-.008) rectangle (.027,.064);
   \draw[psky!75!black,line width=.55pt] (-.014,0)--(-.021,-.07) (.014,0)--(.021,-.07);
  \end{scope}
 }
 \draw[pglow,opacity=.35,line width=.3pt] (.10,.077)--(.32,.077);
 \draw[pglow,opacity=.22,line width=.3pt] (.68,.075)--(.9,.075);
 \end{scope}
 \draw[white,opacity=.35,line width=.25pt,rounded corners=.035cm] (0,0) rectangle (1,1.4);
\end{scope}}

\title{Qwen-Image-PE: Learning Visual Design for\\Text-to-Image Prompt Enhancement}
\author{Zekai Zhang, Yanran Zhang, Shengming Yin, Yan Shu, Zhening Liu,\\
Bingxuan Zhao, Chenyang Zhu, Donghao Zhou, Jiahao Li, Jiahao Wang, Jie Zhang,\\
Kaiyuan Gao, Kun Yan, Lihan Jiang, Ningyuan Tang, Qihang Ai, Rui Song,\\
Shuo Cao, Xiao Xu, Xiaoyue Chen, Yuxiang Chen, Zhendong Wang, Zhouyang Li,\\
Zihao Liu, Zikai Zhou, Bo Zheng, Huishuai Zhang, Dongyan Zhao, Chenfei Wu\thanks{Corresponding author.}}

\begin{document}
\maketitle
\enlargethispage{2\baselineskip}
\vspace{-26pt}
\begin{abstract}
Learning visual design is essential for prompt enhancement (PE) to turn underspecified user requests into coherent visual specifications for text-to-image (T2I) generation. Existing methods preserve user intent and improve overall visual appeal, with less explicit emphasis on fine-grained design dimensions. To address this gap, we introduce Qwen-Image-PE, which learns visual design through two-stage training. Supervised fine-tuning with Caption-Based Reverse Distillation (CRD) uses dense captions of well-designed images as design supervision. Reinforcement learning with Groupwise Design Reward (GDR) jointly compares all candidate images for the same request to refine lighting, composition, layout, and detail. On Qwen-Image-Bench with Qwen-Image-2.1, Qwen-Image-PE-397B improves over original requests by 7.81 points, while Qwen-Image-PE-9B outperforms all evaluated PE baselines. Further evaluations show gains in visual design, aesthetics, image quality, and human preference. We find that PE scales along three axes: training data, model size, and information density, with sufficient information density as a prerequisite for effective enhancement. Moreover, we demonstrate that Qwen-Image-PE generalizes to four additional T2I backbones, even surpassing native prompt enhancers on HunyuanImage-2.1 and FLUX 3 Image Klein. We release Qwen-Image-PE-9B at \url{https://huggingface.co/Qwen/Qwen-Image-2.1-PE-T2I}.
\end{abstract}

\par\vspace{-3pt}\noindent\hfill
\begin{minipage}{\dimexpr\linewidth-2\leftmargini\relax}
\centering
\includegraphics[width=\linewidth]{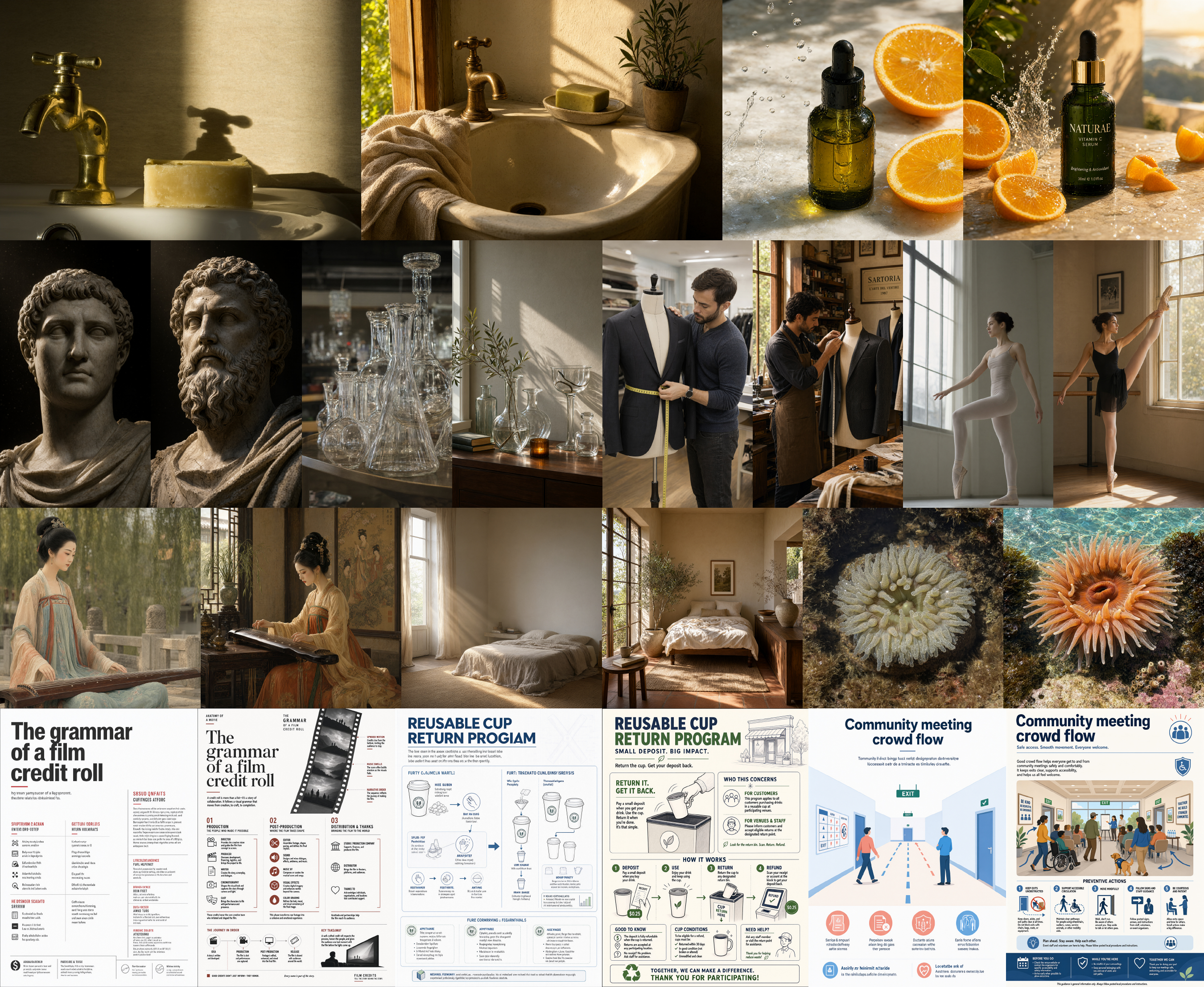}
\captionsetup{font=small,skip=3pt}
\captionof{figure}{\textbf{Qwen-Image-PE improves visual design.} Within each pair, Qwen-Image-2.1 outputs from enhanced prompts (right) show more expressive lighting, more cohesive compositions, clearer layouts, and richer details than those from original requests (left).}
\label{fig:banner}
\end{minipage}\hfill\null\par

\clearpage
\section{Introduction}
\label{sec:intro}

Modern text-to-image (T2I) models can follow increasingly detailed visual specifications, offering control over lighting, composition, layout, and fine-grained detail~\citep{Wu2025QwenImageTR,Zhao2026QwenImage20TR,blackforestlabs2026flux3image}. However, user requests are often brief and leave many of these choices unspecified~\citep{qwenimageagent}. Prompt enhancement (PE) bridges this gap by expanding user requests into detailed prompts. The visual choices introduced during enhancement shape how the requested content is presented, making visual design an important capability for prompt enhancers.

Existing prompt enhancement methods pursue semantic alignment by preserving and clarifying user intent~\citep{promptenhancer,booguimage,ernieimage,blackforestlabs2026flux3image}. Beyond semantic correctness, other approaches improve visual appeal through aesthetic or preference optimization~\citep{promptist,beautifulprompt,prism2026,poet}. However, these approaches primarily optimize overall visual appeal, with less explicit emphasis on individual design dimensions. We therefore study how PE can explicitly learn and refine fine-grained design dimensions, including lighting, composition, layout, and detail. The challenge is how to translate fine-grained visual design into effective supervision targets and optimization feedback.
\looseness=-1 

Our key insight is that: (1) Well-designed images demonstrate how design choices work together. Dense captions make these choices explicit, providing concrete supervision for PE to learn visual design. (2) Jointly comparing all candidate images for the same request provides a common basis for calibrating design scores. This groupwise comparison helps distinguish their relative quality, providing more informative feedback.

Guided by these insights, we introduce Qwen-Image-PE, which learns visual design through a two-stage training paradigm. First, supervised fine-tuning via \textbf{Caption-Based Reverse Distillation (CRD)} enables PE to imitate the design choices in well-designed images. We construct request--image pairs, select images for design quality and request consistency, and densely caption their visual content and design choices to obtain enhanced-prompt targets. These captions, together with synthesized design rationales, provide supervision for learning visual design. Second, reinforcement learning with \textbf{Groupwise Design Reward (GDR)} refines the enhancer through visual comparison. For each request, we generate images from a group of sampled enhanced prompts and arrange the images into a grid. A multimodal judge jointly scores the candidates under explicit design rubrics, conditioning each candidate's scores on the full group. We combine these design scores with an alignment reward to optimize the enhancer while keeping the T2I backbone frozen.

We evaluate Qwen-Image-PE on Qwen-Image-Bench~\citep{li2026qwenimagebench} for overall generation performance, complemented by assessments of visual design, aesthetics, image quality, and human preference. On Qwen-Image-Bench with a common Qwen-Image-2.1 backbone, Qwen-Image-PE-397B improves the Overall score over original requests by 7.81 points, while Qwen-Image-PE-9B outperforms all evaluated prompt enhancement baselines in Overall. Complementary evaluations show gains in visual design, aesthetics, image quality, and human preference.

We further study the scaling properties and generalization behavior of PE, with four main findings. (1) PE scales with training data, model size, and information density. (2) Sufficient information density is a prerequisite for effective enhancement. (3) Both PE variants outperform original requests on four additional T2I backbones without access to their training captions. (4) On HunyuanImage-2.1 and FLUX 3 Image Klein, both variants outperform native prompt enhancers, suggesting that PE learns transferable visual design principles rather than generator-specific caption patterns.

\noindent Our contributions are summarized as follows:
\begin{itemize}[leftmargin=*, itemsep=2pt, topsep=2pt, parsep=0pt]
\item \textbf{Training paradigm of PE.} We propose a two-stage training paradigm for learning visual design. First, supervised fine-tuning via Caption-Based Reverse Distillation (CRD) enables PE to imitate the design choices in well-designed images. Second, reinforcement learning with Groupwise Design Reward (GDR) refines fine-grained design dimensions, including lighting, composition, layout, and detail, through groupwise visual comparison.
\item \textbf{Scaling properties of PE.} We discover that PE performance scales along three axes: training data, model size, and information density. Our experiments suggest that sufficient information density is a prerequisite for effective prompt enhancement in the evaluated setting.
\item \textbf{Generalization behavior of PE.} We demonstrate transfer across four T2I backbones without access to their training captions, with higher Overall scores than native prompt enhancers on HunyuanImage-2.1 and FLUX 3 Image Klein. This suggests that PE learns transferable visual design principles rather than generator-specific caption patterns.
\end{itemize}

\section{Method}
\label{sec:method}

\begin{figure}[t]
\centering
\input{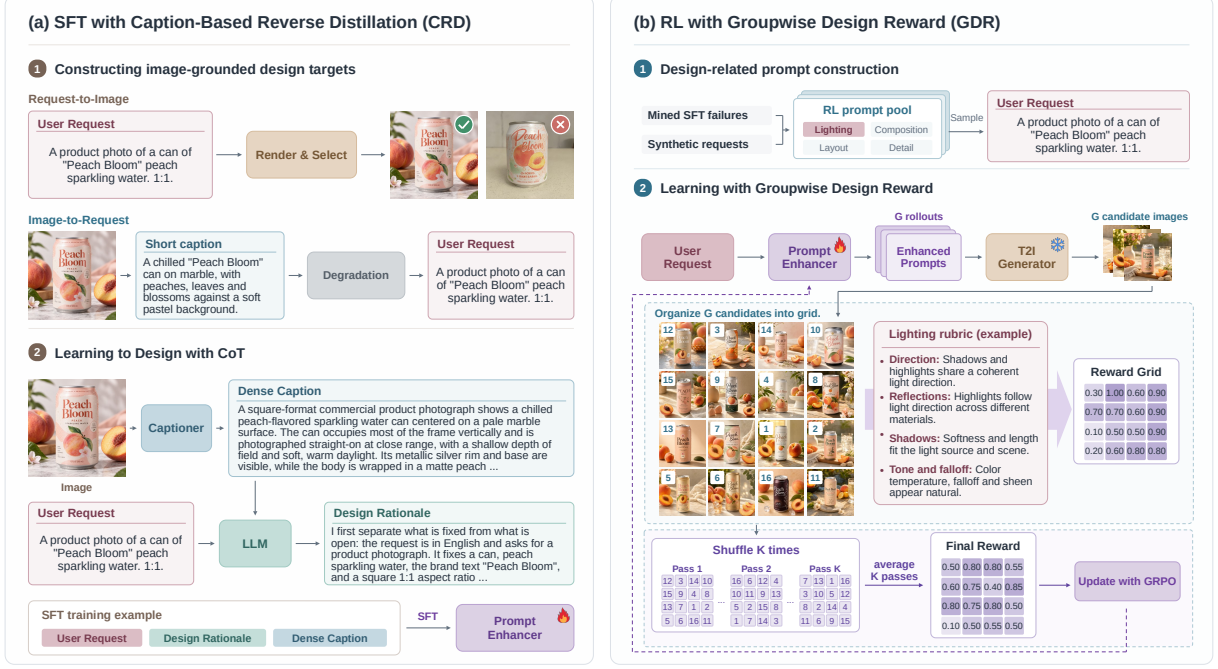}
\caption{\textbf{Overview of Qwen-Image-PE.} \textbf{(a)} Supervised fine-tuning with Caption-Based Reverse Distillation (CRD) trains the enhancer to imitate design choices described in dense captions of well-designed images. \textbf{(b)} Reinforcement learning with Groupwise Design Reward (GDR) refines lighting, composition, layout, and detail by jointly scoring images for the same request, using the full rollout group.}
\label{fig:method}
\end{figure}

Qwen-Image-PE learns visual design through a two-stage training paradigm (\Cref{fig:method}). After formulating prompt enhancement (\Cref{sec:formulation}), we describe supervised fine-tuning with Caption-Based Reverse Distillation (CRD) to learn design choices from well-designed images (\Cref{sec:vpd}) and reinforcement learning with Groupwise Design Reward (GDR) to refine these choices through groupwise visual comparison (\Cref{sec:rl}).

\subsection{Formulation of Prompt Enhancement}
\label{sec:formulation}

Given a user request $x$, a prompt enhancer $\pi_\theta$ generates a response $o=(z,y,\rho)$ containing a chain-of-thought (CoT) rationale $z$, an enhanced prompt $y$, and an aspect ratio $\rho$. The enhanced prompt and aspect ratio condition a frozen text-to-image (T2I) generator $\mathcal{G}$ to synthesize an image $I$:
\begin{equation}
o=(z,y,\rho)\sim\pi_\theta(\cdot\mid x),
\qquad I\sim\mathcal{G}(\cdot\mid y,\rho).
\end{equation}

User requests often leave visual design choices unspecified. Our objective is to learn a prompt enhancer that specifies these choices coherently while preserving the user's explicit requirements. We focus on four fine-grained visual design dimensions. \textbf{Lighting} concerns physically coherent illumination: consistent light direction, shadows, highlights, and reflections, with natural color temperature, light falloff, and material sheen. \textbf{Composition} concerns subject placement and scale, focal hierarchy, balance, and depth layering. \textbf{Layout} concerns the alignment, spacing, whitespace, and typographic hierarchy of text, graphics, and content modules. \textbf{Detail} concerns the richness of material textures, local structures, environmental details, and meaningful informational content.

\Needspace{8\baselineskip}
\subsection{Supervised Fine-Tuning with Caption-Based Reverse Distillation}
\label{sec:vpd}

To ground design supervision in visual examples, we introduce \textbf{Caption-Based Reverse Distillation (CRD)}. Unlike forward-based prompt enhancement, which directly expands user requests into detailed prompts using a zero-shot or trained language model, CRD derives enhanced-prompt targets by densely captioning well-designed images. CRD constructs supervision in the image-to-text direction, reversing the direction of T2I generation. We then train the enhancer to predict these targets from user requests.

\begin{figure}[t]
\centering
\includegraphics[width=\linewidth]{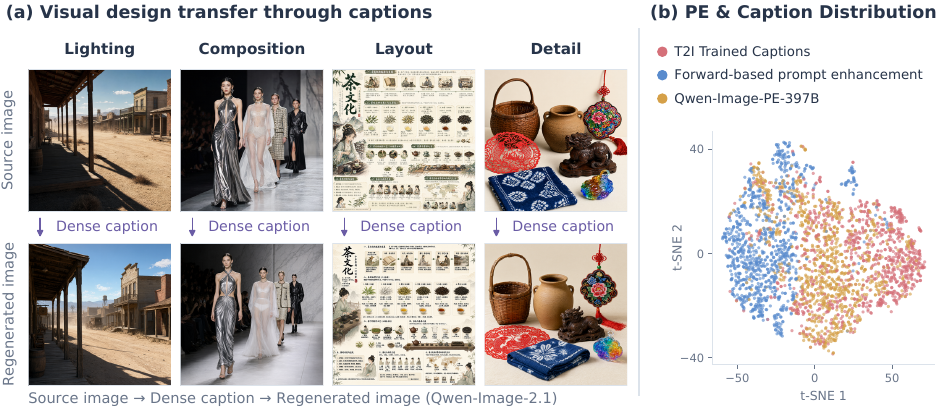}
\caption{\textbf{Motivating Caption-Based Reverse Distillation.} \textbf{(a)} Dense captions of well-designed images guide Qwen-Image-2.1 to reproduce key visual design choices in lighting, composition, layout, and detail. \textbf{(b)} In the t-SNE projection, Qwen-Image-PE-397B prompt embeddings overlap more with embeddings of sampled T2I training captions than forward-based prompt embeddings do.}
\label{fig:vpd_motivation}
\end{figure}

\paragraph{Motivation.}
\textbf{(1) Transfer visual design through captions.} Dense captions describe how visual content is organized and presented in an image. As shown in \Cref{fig:vpd_motivation}a, captions of well-designed images guide Qwen-Image-2.1 to regenerate key aspects of lighting, composition, layout, and detail. These examples suggest that dense captions encode visual design choices in a form that guides T2I generation. This motivates training the enhancer to predict such captions from user requests, thereby learning to imitate the design choices embodied in the source images. \textbf{(2) Bridge the prompt--caption distribution gap.} T2I generators are trained with caption-image pairs, while forward-based prompt enhancement can produce text that differs from their training caption distributions. We obtain enhanced prompts from Qwen-Image-PE-397B and Qwen-3.8-Max for the same 1,000 requests from Qwen-Image-Bench. Using Qwen-Image-2.1 text-encoder embeddings, we jointly visualize these prompts and 1,000 sampled T2I training captions with t-SNE (\Cref{fig:vpd_motivation}b). Our prompt embeddings exhibit greater overlap with the training-caption embeddings than forward-enhanced prompt embeddings do. This suggests that CRD helps narrow the prompt--caption distribution gap by training the enhancer on dense image captions. Further quantitative results in \Cref{sec:caption_proximity} support this closer alignment.
\looseness=-1 

\paragraph{Constructing image-grounded design targets.}
We construct and select request--image pairs $(x_i,I_i)$ through two complementary paths. In \emph{Request-to-Image} construction, a strong T2I system generates images from real and synthetic requests; we select images for design quality and request consistency. In \emph{Image-to-Request} construction, we start from quality-filtered images and degrade their short captions into noisier, less structured user requests, preserving the core intent while omitting design details.

For each selected image $I_i$, a multimodal captioner produces a dense description $y_i^\star$ that serves as the enhanced-prompt target. The caption describes the image's visual content and design choices across lighting, composition, layout, and detail. We also record the image's aspect ratio $\rho_i^\star$:
\begin{equation}
y_i^\star=\operatorname{Captioner}(I_i),\qquad
\rho_i^\star=\frac{\operatorname{width}(I_i)}{\operatorname{height}(I_i)}.
\label{eq:caption}
\end{equation}

\Needspace{8\baselineskip}
\paragraph{Learning to design with CoT.}
To connect the user's requirements to the image-grounded design target, we augment each caption with a CoT rationale. Given the request $x_i$ and target caption $y_i^\star$, a language model synthesizes
\begin{equation}
z_i^\star=\operatorname{LLM}(x_i,y_i^\star).
\label{eq:cot_synthesis}
\end{equation}
The rationale identifies requirements to preserve and explains the design choices expressed in the target caption. Rationale synthesis uses the target caption as context, while the enhancer learns to predict the rationale from the user request. We collect the annotated examples into
\begin{equation}
\mathcal{D}_{\mathrm{SFT}}
=\left\{(x_i,z_i^\star,y_i^\star,\rho_i^\star)\right\}_{i=1}^{N},
\label{eq:sft_dataset}
\end{equation}
where each tuple contains a user request, a synthesized CoT rationale, an enhanced-prompt target, and an aspect ratio. We train the enhancer to generate the CoT rationale followed by the enhanced prompt and aspect ratio, minimizing
\begin{equation}
\mathcal{L}_{\mathrm{SFT}}(\theta)=
-\frac{1}{N}\sum_{i=1}^{N}\Big[
\log\pi_\theta(z_i^\star\mid x_i)
+\log\pi_\theta(y_i^\star,\rho_i^\star\mid x_i,z_i^\star)
\Big].
\label{eq:sft}
\end{equation}
This objective trains PE to generate design rationales and enhanced prompts that express the design choices demonstrated by the selected images.

\Needspace{8\baselineskip}
\subsection{Reinforcement Learning with Groupwise Design Reward}
\label{sec:rl}

To further refine visual design, we introduce \textbf{Groupwise Design Reward (GDR)}. Existing reward models, such as PickScore~\citep{pickapic} and HPSv3~\citep{ma2025hpsv3}, provide overall preference scores without explicit feedback on our four fine-grained design dimensions. Given the strong visual understanding and reasoning capabilities of MLLMs, we use an MLLM-based judge to assess these dimensions under explicit rubrics. While pointwise judging can assess individual images, it lacks shared visual references for distinguishing subtle design differences. Pairwise judging provides relative comparisons, but lacks a holistic view of the rollout group and incurs substantial evaluation overhead when all candidate pairs are compared. GDR combines full rollout-group context with efficient joint scoring. By evaluating all candidates together, it provides shared visual references while avoiding separate judge calls for individual images or candidate pairs.

\paragraph{Design-related prompt construction.}
We construct the RL prompt pool $\mathcal{D}_{\mathrm{RL}}$ from mined and synthetic requests. We first collect requests for which prompts from the SFT enhancer yield images with weaknesses in visual design. We then synthesize additional requests to broaden the coverage of design tasks and requirements.

\paragraph{Learning with groupwise design rewards.}
For each request $x\sim\mathcal{D}_{\mathrm{RL}}$, the enhancer samples $G$ responses, each containing a CoT rationale $z_i$, an enhanced prompt $y_i$, and an aspect ratio $\rho_i$. The frozen T2I generator renders an image for each response:
\begin{equation}
o_i=(z_i,y_i,\rho_i)\sim\pi_\theta(\cdot\mid x),
\qquad I_i\sim\mathcal{G}(\cdot\mid y_i,\rho_i),
\quad i=1,\ldots,G.
\label{eq:rl_rollout}
\end{equation}
We collect the resulting images into a candidate group $\mathcal{I}_x=(I_1,\ldots,I_G)$, using a shared generator sampling seed within the group.

Let $d\in\{1,\ldots,D\}$ index the design dimensions and $q_d$ denote the corresponding evaluation rubric. GDR presents all candidates for the same request together in a grid and conditions each score on the full rollout group. For each dimension, a multimodal judge jointly scores all candidates in a single call, assessing their relative design quality under the same request and rubric.

To reduce sensitivity to presentation order, we evaluate $K$ shuffled grids. For pass $k$, let $\operatorname{Shuffle}_k(\mathcal{I}_x)$ denote the grid formed by the $k$-th random permutation of the candidate images. We normalize the returned scores to $[0,1]$ and reorder them to match the original candidate order:
\begin{equation}
\left(r_{1,d}^{\mathrm{design},(k)},\ldots,r_{G,d}^{\mathrm{design},(k)}\right)
=\operatorname{Judge}\!\left(x,\operatorname{Shuffle}_k(\mathcal{I}_x),q_d\right),
\label{eq:reward_context}
\end{equation}
where $\operatorname{Judge}$ denotes joint evaluation followed by score normalization and candidate matching. The final design reward averages the scores assigned to the same image across presentations:
\begin{equation}
r_{i,d}^{\mathrm{design}}
=\frac{1}{K}\sum_{k=1}^{K}r_{i,d}^{\mathrm{design},(k)}.
\label{eq:judge}
\end{equation}
For $D$ dimensions, GDR uses $DK$ judge calls per group, compared with $DG$ for single-pass pointwise scoring and $D\binom{G}{2}$ for exhaustive pairwise scoring. GDR requires fewer calls than both alternatives as we adopt $K\ll G$.

We combine the groupwise design rewards with an image--request alignment reward:
\begin{equation}
R_i=\sum_{d=1}^{D}\lambda_d r_{i,d}^{\mathrm{design}}
+\lambda_{\mathrm{a}}r_i^{\mathrm{align}},
\label{eq:reward}
\end{equation}
where $\lambda_d$ weights the reward for design dimension $d$, and $\lambda_{\mathrm{a}}$ weights the alignment reward. The alignment term encourages preservation of the user's explicit requirements during design optimization.

\paragraph{Policy optimization with GRPO.}
We initialize the enhancer with the policy learned through Caption-Based Reverse Distillation and optimize it with Group Relative Policy Optimization (GRPO)~\citep{shao2024grpo}. The total rewards are normalized within each group to compute relative advantages for updating the enhancer. We apply Kullback--Leibler (KL) regularization toward a frozen copy of the initial SFT policy and keep the T2I generator frozen.

\section{Experiments}
\label{sec:experiments}

We first evaluate whether Qwen-Image-PE improves generation quality and visual design. After describing the experimental setup (\Cref{sec:exp_setup}), we present quantitative results, human assessment, and qualitative comparisons (\Cref{sec:exp_quant}). We then examine the contributions of Caption-Based Reverse Distillation (CRD) and Groupwise Design Reward (GDR) (\Cref{sec:exp_ablation}). Furthermore, we study the scaling properties and generalization in \Cref{sec:scaling,sec:generalization}.

\subsection{Experimental Setup}
\label{sec:exp_setup}

\paragraph{Models.}
We develop Qwen-Image-PE-9B and Qwen-Image-PE-397B, initialized from Qwen3.5-9B and Qwen3.5-397B-A17B~\citep{qwen3.5}, respectively. We train both variants through supervised fine-tuning with Caption-Based Reverse Distillation (CRD), followed by reinforcement learning with Groupwise Design Reward (GDR), while keeping the T2I backbone frozen.

\begin{table}[t]
\centering
\small
\setlength{\tabcolsep}{5.5pt}
\renewcommand{\arraystretch}{1.12}
\resizebox{\linewidth}{!}{%
\begin{tabular}{lcccccc}
\toprule
\multirow{3}{*}{\textbf{Model}} & \multicolumn{5}{c}{\textbf{Evaluation Dimension}} & \multirow{3}{*}{\textbf{Overall}} \\
\cmidrule(lr){2-6}
& \multirow{2}{*}{\textit{Quality}} & \multirow{2}{*}{\textit{Aesthetics}} & \multirow{2}{*}{\textit{Alignment}} & \textit{Real-world} & \textit{Creative} & \\
& & & & \textit{Fidelity} & \textit{Generation} & \\
\midrule
GPT Image 2 & 58.65 & 67.53 & 65.85 & 57.38 & 75.23 & 64.69 \\
Nano Banana 2.0 & 54.77 & 61.08 & 62.40 & 54.28 & 67.05 & 59.82 \\
GPT Image 1.5 & 55.14 & 60.88 & 61.72 & 53.95 & 66.35 & 59.65 \\
Nano Banana Pro & 55.67 & 60.26 & 61.25 & 54.07 & 66.23 & 59.45 \\
Qwen Image 2.0 Pro & 54.39 & 58.67 & 59.28 & 51.83 & 64.94 & 57.84 \\
Seedream 5.0 & 52.55 & 58.40 & 58.90 & 51.92 & 65.29 & 57.22 \\
FLUX 2 Max & 53.64 & 56.85 & 57.35 & 49.35 & 56.50 & 55.33 \\
HunyuanImage 3.0 & 50.35 & 53.57 & 52.00 & 44.31 & 49.12 & 50.81 \\
\midrule
\rowcolor{blue!8}
\multicolumn{7}{c}{T2I backbone: Qwen-Image-2.1} \\
\rowcolor{pegray!9}
\multicolumn{7}{c}{\footnotesize\textit{w/o Prompt Enhancer}} \\
Original Request & 55.09 & 54.47 & 54.59 & 51.38 & 54.47 & 54.36 \\
\rowcolor{pegray!9}
\multicolumn{7}{c}{\footnotesize\textit{Closed-Source API-Based Prompt Enhancers}} \\
Qwen-3.8-Max & 55.83 & 60.85 & 61.31 & 53.74 & 66.26 & 59.59 \\
Gemini-3.1-Pro & 56.49 & 61.31 & 61.04 & 55.04 & 66.61 & 60.04 \\
GPT-5.5 & 55.62 & 61.60 & 62.11 & 54.55 & 67.25 & 60.10 \\
\rowcolor{pegray!9}
\multicolumn{7}{c}{\footnotesize\textit{Open-Source Trained Prompt Enhancers}} \\
HunyuanImage-2.1-PE-7B & 55.02 & 57.35 & 56.59 & 51.05 & 57.90 & 55.96 \\
HunyuanImage-2.1-PE-32B & 54.19 & 58.98 & 57.55 & 53.30 & 61.45 & 57.13 \\
ERNIE-Image-PE & 55.78 & 60.28 & 59.28 & 53.85 & 64.53 & 58.70 \\
FLUX 3 Image Klein PE & 55.61 & 60.84 & 61.86 & 54.44 & 67.14 & 59.90 \\
\rowcolor{blue!8}
\textbf{Qwen-Image-PE-9B} & 57.28 & 62.48 & 60.72 & 54.47 & 66.36 & 60.28 \\
\rowcolor{blue!8}
\textbf{Qwen-Image-PE-397B}$^{\dagger}$ & \textbf{57.95} & \textbf{64.50} & \textbf{62.63} & \textbf{55.57} & \textbf{70.89} & \textbf{62.17} \\
\bottomrule
\end{tabular}%
}
\caption{\textbf{Results on Qwen-Image-Bench.} Prompt enhancers share a frozen Qwen-Image-2.1 backbone. Bold marks the best score among methods using this backbone. $^{\dagger}$ marks our internal model. Qwen-Image-PE-397B leads across all five dimensions, while Qwen-Image-PE-9B surpasses all PE baselines.}
\label{tab:qib}
\end{table}

\paragraph{Baselines.}
We compare Qwen-Image-PE with three groups of baselines: (1) \textbf{Original Request}, which uses the user's request directly without enhancement; (2) \textbf{Closed-Source API-Based Prompt Enhancers}, including Qwen-3.8-Max~\citep{qwen38}, Gemini-3.1-Pro~\citep{gemini31pro}, and GPT-5.5~\citep{gpt55systemcard}, which we access through APIs for zero-shot prompt enhancement using the same system prompt as Qwen-Image-PE; and (3) \textbf{Open-Source Trained Prompt Enhancers}, including HunyuanImage-2.1-PE-7B, HunyuanImage-2.1-PE-32B~\citep{promptenhancer}, ERNIE-Image-PE~\citep{ernieimage}, and FLUX 3 Image Klein PE~\citep{blackforestlabs2026flux3image}, using their native prompting interfaces. Unless otherwise stated, all prompt-enhancement comparisons use Qwen-Image-2.1~\citep{Zhao2026QwenImage20TR} as the T2I backbone, with images generated at 2K resolution and matched per-request sampling seeds.

\Needspace{6\baselineskip}
\paragraph{Evaluation.}
We assess overall generation performance, visual design, aesthetics, image quality, and human preference. The requests used for model evaluation do not overlap with those used for SFT or RL. The evaluation metrics described below are distinct from the training reward functions and are not used to optimize the enhancer. \textbf{Qwen-Image-Bench}~\citep{li2026qwenimagebench} covers real-world creation scenarios using criteria developed with professional artists. Its hierarchy comprises five L1 dimensions (Quality, Aesthetics, Alignment, Real-world Fidelity, and Creative Generation) and finer L2 capabilities, including design-related dimensions like Lighting, Composition, Layout, and Detail. On images generated from Qwen-Image-Bench requests, \textbf{UniPercept}~\citep{cao2025unipercept} provides Image Aesthetics Assessment (IAA), Image Quality Assessment (IQA), and Image Structure and Texture Assessment (ISTA), measuring aesthetic appeal, visual quality, and structural and textural richness. We further assess aesthetics-related human preference using \textbf{HPSv3} and \textbf{PickScore}~\citep{pickapic} on images generated from GenEval prompts~\citep{Ghosh2023GenEvalAO}, whose shorter length accommodates these metrics' text-input limits. Finally, \textbf{Good/Same/Bad (GSB)} human evaluation assesses request fulfillment and visual appeal, recording whether annotators prefer our output, consider the outputs comparable, or prefer the baseline.\looseness=-1

\subsection{Main Results}
\label{sec:exp_quant}

\paragraph{State-of-the-art prompt enhancement on Qwen-Image-Bench.}
\Cref{tab:qib} compares original requests with API-based and trained prompt enhancers using the same Qwen-Image-2.1 backbone. Qwen-Image-PE-397B achieves the highest Overall score of 62.17 among the evaluated enhancers, improving over Original Request (54.36) by 7.81 points and leading across all five dimensions. Qwen-Image-PE-9B outperforms all PE baselines in Overall, scoring 60.28 compared with 60.10 for GPT-5.5 and 59.90 for FLUX 3 Image Klein PE. These results show that Qwen-Image-PE improves overall generation performance.

\paragraph{Improved visual design and aesthetics.}
\Cref{tab:additional_eval} evaluates fine-grained visual design, aesthetics, and image quality. Qwen-Image-PE-397B leads in Lighting, Composition, and Detail, while its Layout score closely matches that of GPT-5.5 (61.11 vs.\ 61.15). Qwen-Image-PE-9B outperforms all trained baselines across the four design dimensions. Both variants also surpass all evaluated baselines on the three UniPercept metrics, HPSv3, and PickScore, with the 397B model consistently ranking first. Together, these results support improvements in visual design capabilities, image aesthetics, and quality.

\begin{table}[t]
\centering
\small
\setlength{\tabcolsep}{3.2pt}
\renewcommand{\arraystretch}{1.12}
\resizebox{\linewidth}{!}{%
\begin{tabular}{l*{9}{c}}
\toprule
\multirow{2}{*}{\textbf{Model}} & \multicolumn{4}{c}{\textbf{QIB Design-related Capabilities}} & \multicolumn{3}{c}{\textbf{UniPercept}} & \multirow{2}{*}{\textbf{HPSv3} $\uparrow$} & \multirow{2}{*}{\textbf{PickScore} $\uparrow$} \\
\cmidrule(lr){2-5}\cmidrule(lr){6-8}
& \textit{Lighting} & \textit{Composition} & \textit{Layout} & \textit{Detail} & \textit{IAA} & \textit{IQA} & \textit{ISTA} & & \\
\midrule
\rowcolor{pegray!9}
\multicolumn{10}{c}{\footnotesize\textit{w/o Prompt Enhancer}} \\
Original Request & 64.12 & 54.68 & 54.88 & 55.15 & 66.28 & 72.21 & 43.71 & 6.3773 & 22.2699 \\
\rowcolor{pegray!9}
\multicolumn{10}{c}{\footnotesize\textit{Closed-Source API-Based Prompt Enhancers}} \\
Qwen-3.8-Max & 71.02 & 62.68 & 60.34 & 56.73 & 70.17 & 74.34 & 46.64 & 8.5632 & 23.2987 \\
Gemini-3.1-Pro & 70.81 & 61.78 & 60.16 & 57.23 & 70.38 & 74.05 & 45.49 & 9.4122 & 23.5107 \\
GPT-5.5 & 71.50 & \underline{63.88} & \textbf{61.15} & 57.00 & 70.85 & 74.66 & 48.53 & 9.1513 & 23.4492 \\
\rowcolor{pegray!9}
\multicolumn{10}{c}{\footnotesize\textit{Open-Source Trained Prompt Enhancers}} \\
HunyuanImage-2.1-PE-7B & 66.84 & 57.40 & 56.04 & 55.36 & 67.94 & 73.10 & 43.67 & 7.6252 & 23.0570 \\
HunyuanImage-2.1-PE-32B & 69.18 & 59.26 & 56.85 & 54.65 & 68.64 & 72.78 & 44.34 & 8.2167 & 23.2575 \\
ERNIE-Image-PE & 70.15 & 60.40 & 58.73 & 56.83 & 69.52 & 74.65 & 46.47 & 8.8924 & 23.4778 \\
FLUX 3 Image Klein PE & 70.11 & 62.94 & 60.31 & 56.62 & 70.01 & 74.37 & 49.16 & 8.9679 & 23.5169 \\
\rowcolor{blue!8}
\textbf{Qwen-Image-PE-9B} & \underline{75.51} & 63.80 & 60.83 & \underline{58.18} & \underline{72.82} & \underline{76.42} & \underline{54.10} & \underline{9.8670} & \underline{23.5343} \\
\rowcolor{blue!8}
\textbf{Qwen-Image-PE-397B}$^{\dagger}$ & \textbf{76.59} & \textbf{66.66} & \underline{61.11} & \textbf{58.91} & \textbf{73.73} & \textbf{76.65} & \textbf{55.12} & \textbf{9.9553} & \textbf{23.7541} \\
\bottomrule
\end{tabular}%
}
\caption{\textbf{Evaluation of visual design and aesthetics.} Bold and underlined scores indicate the best and second-best results per column, respectively. $^{\dagger}$ marks our internal model. Qwen-Image-PE-397B ranks first on eight of nine metrics, while Qwen-Image-PE-9B outperforms trained PE baselines on all metrics.}
\label{tab:additional_eval}
\end{table}

\par\medskip
\noindent
\begin{minipage}[b]{0.48\textwidth}
\textbf{Visual design aligns with human preferences.}\par
To assess human preference for the trained enhancer's outputs, we conduct GSB evaluation with expert annotators on 500 cases balanced across domains. In \Cref{fig:gsb_preview}, Qwen-Image-PE-397B achieves positive net win rates, $(G-B)/(G+S+B)$, against all baselines: 35.48\% over Original Request, 32.30\% over HunyuanImage-2.1-PE-32B, 22.49\% over ERNIE-Image-PE, 17.09\% over GPT-5.5, and 13.60\% over FLUX 3 Image Klein PE. These results indicate that annotators prefer images generated with Qwen-Image-PE-397B over baseline outputs.
\end{minipage}\hfill
\begin{minipage}[b]{0.50\textwidth}
\centering
\includegraphics[width=\linewidth]{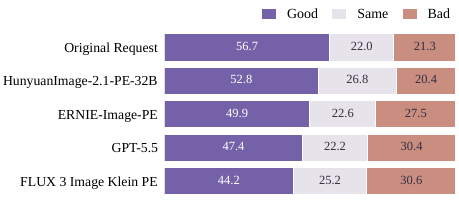}
\captionsetup{font=footnotesize,skip=5pt}
\captionof{figure}{\textbf{GSB evaluation results.} Each row compares Qwen-Image-PE-397B with a baseline. Percentages indicate preference for our output (Good), comparable outputs (Same), or preference for the baseline (Bad).}
\label{fig:gsb_preview}
\vspace{-3.3pt}
\end{minipage}

\begin{figure}[t]
\centering
\includegraphics[width=\linewidth]{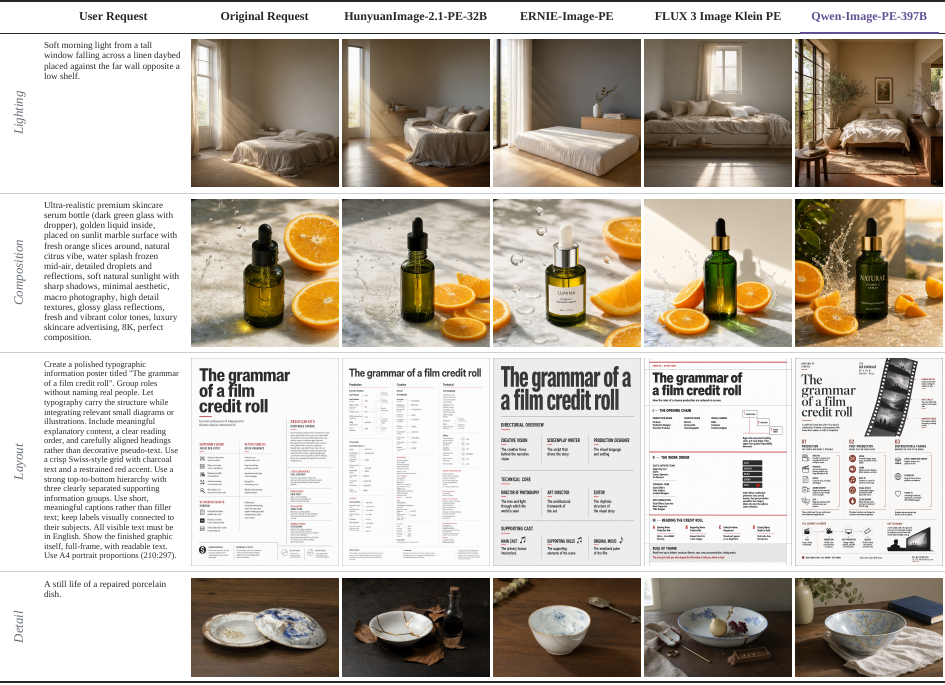}
\captionsetup{font=small,skip=3pt}
\caption{\textbf{Qualitative comparisons across four design dimensions.} In these examples, prompts from Qwen-Image-PE-397B yield more expressive lighting, more cohesive compositions, better-organized layouts, and richer details than original requests or prompts from baseline enhancers.}
\label{fig:qualitative_comparison}
\end{figure}

\paragraph{Qualitative comparisons.}
\label{sec:exp_qual}
\Cref{fig:qualitative_comparison} compares images generated using original requests, baseline enhancers, and Qwen-Image-PE-397B across four design dimensions. In the Lighting example, directional light and cast shadows emphasize the bed's form and the folds of the bedding. In the Composition example, the bottle, citrus, and water splash occupy different depths, while the bottle remains the visual focus. In the Layout example, aligned content groups and numbered headings establish a clear hierarchy and reading order. In the Detail example, repair lines follow the bowl's curvature, while contrasting textures distinguish the porcelain from the fabric beneath it. These examples show how Qwen-Image-PE translates user requests into coordinated visual design choices.

\FloatBarrier
\subsection{Ablation Studies}
\label{sec:exp_ablation}

We examine the two training stages in \Cref{sec:method}: whether supervised fine-tuning with Caption-Based Reverse Distillation helps the model learn visual design, and whether reinforcement learning with Groupwise Design Reward further refines these design choices.

\begin{figure}[!t]
\centering
\includegraphics[width=\linewidth]{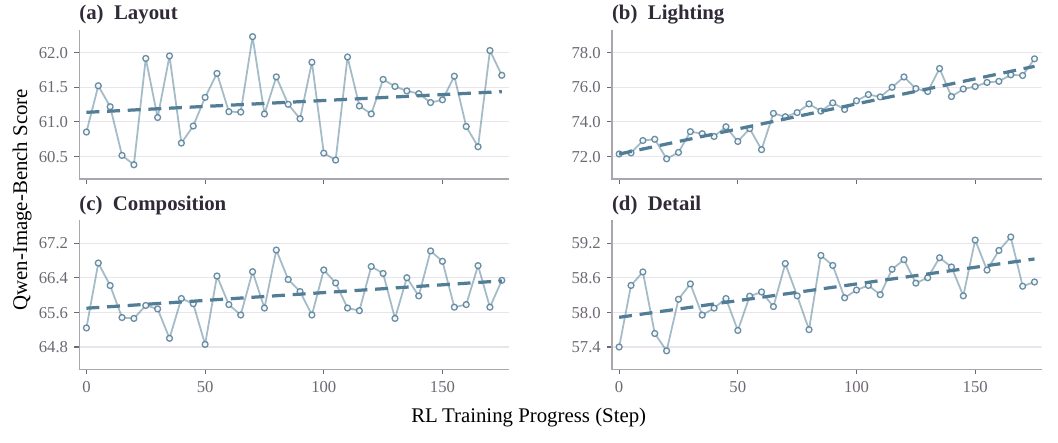}
\caption{\textbf{Benchmark performance during RL training.} The curves show benchmark evaluation scores, not training rewards. Qwen-Image-PE-397B improves across design dimensions on Qwen-Image-Bench, with the most consistent gains in Lighting. Dashed lines indicate linear trends.}
\label{fig:qib_rl_trace}
\end{figure}

\begin{table}[htbp]
\centering
\small
\setlength{\tabcolsep}{6pt}
\renewcommand{\arraystretch}{1.08}
\resizebox{\linewidth}{!}{%
\begin{tabular}{lcccccc}
\toprule
\multirow{2}{*}{Method} & \multirow{2}{*}{Quality} & \multirow{2}{*}{Aesthetics} & \multirow{2}{*}{Alignment} & Real-world & Creative & \multirow{2}{*}{Overall} \\
& & & & Fidelity & Generation & \\
\midrule
\rowcolor{blue!8}
\multicolumn{7}{c}{\textbf{(a) Supervised Fine-Tuning}} \\
\rowcolor{pegray!9}
\multicolumn{7}{c}{\textit{T2I backbone: Qwen-Image-2.1}} \\
Original Request & 55.09 & 54.47 & 54.59 & 51.38 & 54.47 & 54.36 \\
+ Forward-Based Supervision & 55.98 & 60.34 & 61.77 & 54.00 & 67.70 & 59.86 \\
+ Caption-Based Reverse Distillation & 57.10 & 63.34 & 62.33 & 55.82 & 70.12 & 61.52 \\
\midrule
\rowcolor{blue!8}
\multicolumn{7}{c}{\textbf{(b) Reinforcement Learning}} \\
\rowcolor{pegray!9}
\multicolumn{7}{c}{\textit{T2I backbone: Qwen-Image-2.1}} \\
SFT initialization & 57.10 & 63.34 & 62.33 & 55.82 & 70.12 & 61.52 \\
+ Pointwise Design Reward & 56.96 & 62.80 & 61.92 & 56.70 & 70.03 & 61.36 \\
+ Groupwise Design Reward & 57.95 & 64.50 & 62.63 & 55.57 & 70.89 & 62.17 \\
\bottomrule
\end{tabular}%
}
\caption{\textbf{Ablation of the two training stages.} For Qwen-Image-PE-397B, Caption-Based Reverse Distillation outperforms Forward-Based Supervision across all five dimensions. Groupwise Design Reward further improves Overall over both the SFT initialization and Pointwise Design Reward.}
\label{tab:training_ablation}
\end{table}

\Needspace{12\baselineskip}
\subsubsection{Effectiveness of Caption-Based Reverse Distillation}
\label{sec:exp_crd}
\label{sec:exp_vdd}

\paragraph{Learning visual design from image-grounded targets.}
We compare two strategies for supervised fine-tuning of the 397B enhancer, using the same multimodal language model (MLLM) to construct targets and identical training settings. Forward-Based Supervision uses the MLLM's zero-shot enhanced prompts as targets, while Caption-Based Reverse Distillation uses it as a dense captioning model. We evaluate both models on Qwen-Image-Bench using Qwen-Image-2.1 as the T2I backbone. As shown in \Cref{tab:training_ablation}(a), Caption-Based Reverse Distillation improves Overall from 59.86 to 61.52 (+1.66) and achieves higher scores across all five dimensions. The largest gains occur in Aesthetics (+3.00) and Creative Generation (+2.42). These results support caption-based supervision as an effective way to transfer visual design choices from well-designed images to the enhancer.

\Needspace{13\baselineskip}
\begin{wraptable}[9]{r}{0.50\textwidth}
\centering
\captionsetup{font=footnotesize,skip=3pt}
\begingroup
\fontsize{8.5}{10}\selectfont
\setlength{\tabcolsep}{1.8pt}
\renewcommand{\arraystretch}{1.14}
\resizebox{\linewidth}{!}{%
\begin{tabular}{@{}lcccc@{}}
\toprule
\multirow{2}{*}{Method} & \multicolumn{2}{c}{JS $\downarrow$} & \multirow{2}{*}{FD $\downarrow$} & \multirow{2}{*}{MAUVE $\uparrow$} \\
\cmidrule(lr){2-3}
& Unigram & Bigram & & \\
\midrule
Qwen-3.8-Max & 0.229 & 0.539 & 0.0709 & 0.032 \\
\rowcolor{blue!8}
Qwen-Image-PE-9B & 0.111 & 0.347 & 0.0221 & 0.221 \\
\rowcolor{blue!8}
Qwen-Image-PE-397B & \textbf{0.105} & \textbf{0.329} & \textbf{0.0207} & \textbf{0.414} \\
\bottomrule
\end{tabular}%
}
\endgroup
\caption{\textbf{Prompt--caption proximity.} Lower JS and FD and higher MAUVE indicate closer distributions.}
\label{tab:caption_proximity}
\vspace{-8pt}
\end{wraptable}
\paragraph{Closer alignment with caption distribution.}
\label{sec:caption_proximity}
To measure the prompt--caption distribution gap, we compare enhanced prompts generated for Qwen-Image-Bench with an equal number of sampled real training captions. We measure \textbf{token-level} differences using unigram and bigram Jensen--Shannon (JS) divergence over corpus-wide token frequencies, \textbf{distribution-level} similarity using MAUVE~\citep{pillutla2021mauve}, and \textbf{embedding-level} distance using Fr\'echet distance (FD). MAUVE and FD are computed from Qwen-Image-2.1 text-encoder representations. \Cref{tab:caption_proximity} shows that both PE variants achieve lower JS and FD and higher MAUVE than Qwen-3.8-Max. In particular, Qwen-Image-PE-397B reduces unigram/bigram JS from 0.229/0.539 to 0.105/0.329 and increases MAUVE from 0.032 to 0.414. Together with \Cref{fig:vpd_motivation}b, these results show that Qwen-Image-PE produces prompts closer to the training-caption distribution across all three levels, supporting the motivation for CRD.

\subsubsection{Effectiveness of Groupwise Design Reward}
\label{sec:exp_gdr}
\label{sec:gdr_evaluation}

\paragraph{Refining visual design through groupwise feedback.}
Starting from the 397B model trained with Caption-Based Reverse Distillation, we compare reinforcement learning with Pointwise Design Reward and Groupwise Design Reward. Both use the same design rubric: Pointwise Design Reward evaluates each image independently, while Groupwise Design Reward jointly scores candidate images for the same request. We do not include exhaustive pairwise judging, as its quadratic number of MLLM calls per rollout group makes it impractical for online RL at our training scale. We evaluate both models on Qwen-Image-Bench using Qwen-Image-2.1 as the T2I backbone. As shown in \Cref{tab:training_ablation}(b), Groupwise Design Reward achieves 62.17 Overall, exceeding Pointwise Design Reward by 0.81 points and the initialization by 0.65 points. It improves four of the five dimensions over pointwise scoring, with the largest gain in Aesthetics (+1.70). These results support groupwise visual comparison as effective feedback for refining visual design.

\paragraph{Agreement with human design preferences.}
To evaluate the ranking signal used during training, we compare reward orderings with blinded human judgments on rollout image pairs across Lighting, Composition, Layout, and Detail. Agreement measures how often the higher-scoring image matches the annotator's preference. Groupwise Design Reward achieves 59.1\% agreement, compared with 50.9\% for Pointwise Design Reward (+8.2 percentage points). These results suggest that groupwise scoring provides a ranking signal that better reflects human judgments of visual design.

\paragraph{Benchmark performance during RL training.}
We evaluate Qwen-Image-PE-397B on Qwen-Image-Bench during RL training, using Qwen-Image-2.1 as the T2I backbone (\Cref{fig:qib_rl_trace}). Lighting exhibits the most consistent improvement, rising from 72.14 to 77.64. Composition and Detail show positive but noisier trends, while Layout improves only slightly with substantial fluctuations. While GDR provides MLLM-based groupwise training rewards, benchmark evaluation uses pointwise scoring by Q-Judger, a separate evaluator trained on expert annotations. This consistency between GDR's design objectives and independently measured improvements further supports its effectiveness in refining visual design.

\FloatBarrier
\begin{figure}[!t]
\centering
\includegraphics[width=\linewidth]{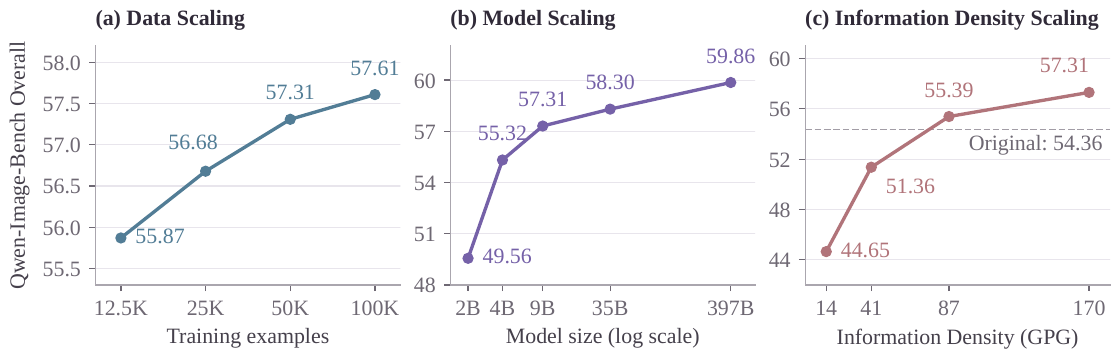}
\caption{\textbf{Scaling properties of PE.} PE scales with \textbf{(a)} training data, \textbf{(b)} model size, and \textbf{(c)} information density. Sufficient information density is a prerequisite for effective prompt enhancement.}
\label{fig:qib_scaling}
\end{figure}

\section{Scaling Properties of Qwen-Image-PE}
\label{sec:scaling}
\label{sec:exp_analysis}

We study the scaling properties of prompt enhancement along three axes: training data, model size and information density. All experiments evaluate SFT models on Qwen-Image-Bench, using \mbox{Qwen-Image-2.1} as the frozen T2I backbone.

\paragraph{Scaling Training Data.}
We sample nested subsets of 12.5K, 25K, 50K, and 100K examples from our training data, such that each larger subset contains all examples in the smaller subsets. We train a \mbox{Qwen3.5-9B} PE model on each subset, retaining the original caption supervision and following the same SFT recipe. As shown in \Cref{fig:qib_scaling}a, Overall increases from 55.87 to 56.68, 57.31, and 57.61 as the training set grows, for a total gain of 1.74 points. Each doubling improves performance, but the incremental gains decrease from 0.81 to 0.63 and then 0.30 points. These results show that additional training data consistently benefits the enhancer, although the smaller gain from 50K to 100K indicates diminishing returns over the evaluated range.

\paragraph{Scaling Model Size.}
We fix the training set to the same 50K examples, retain their original caption supervision, and train five PE models following the same SFT recipe: \mbox{Qwen3.5-2B}, \mbox{Qwen3.5-4B}, \mbox{Qwen3.5-9B}, \mbox{Qwen3.5-35B-A3B}, and \mbox{Qwen3.5-397B-A17B}. The sweep includes both dense and MoE models. Overall increases monotonically from 49.56 for the smallest model to 59.86 for the largest (\Cref{fig:qib_scaling}b), a gain of 10.30 points. Comparing the smallest and largest models, all five benchmark dimensions improve, with the largest gains in Creative Generation (+19.66) and Alignment (+14.80). With the training data fixed, larger enhancers improve both creative generation and adherence to the user's requirements over the evaluated range.

\paragraph{Scaling Information Density.}
We fix the PE model at \mbox{Qwen3.5-9B} and the training-data size at 50K examples, providing captions at four information-density levels (L1--L4) as supervision. We measure the information content of each trained model's sampled enhanced prompts using \textbf{Grounded Perplexity Gain (GPG)}~\citep{structuredprompt}. GPG measures how much observing a paired image increases the text's likelihood under a frozen vision-language model. For each model's outputs on Qwen-Image-Bench, we use frozen Qwen3-VL-8B-Instruct to sum token log-likelihood gains from observing the corresponding generated images and average over prompts. Mean total GPG for L1--L4 is 14.42, 41.16, 87.25, and 170.36.

As shown in \Cref{fig:qib_scaling}c, Overall rises from 44.65 to 51.36, 55.39, and 57.31 across the four levels, a total improvement of 12.66 points. Information density yields the largest Overall gain among the three scaling axes over the evaluated ranges, highlighting its central role in PE performance. Interestingly, L1 and L2 underperform Original Request (54.36), whereas L3 and L4 exceed this baseline by 1.03 and 2.95 points. This transition from degradation to improvement suggests that sufficient information density is a prerequisite for effective prompt enhancement in this setting.

\FloatBarrier
\begin{table}[!t]
\centering
\small
\setlength{\tabcolsep}{6pt}
\renewcommand{\arraystretch}{1.08}
\resizebox{\linewidth}{!}{%
\begin{tabular}{lcccccc}
\toprule
\multirow{2}{*}{Method} & \multirow{2}{*}{Quality} & \multirow{2}{*}{Aesthetics} & \multirow{2}{*}{Alignment} & Real-world & Creative & \multirow{2}{*}{Overall} \\
& & & & Fidelity & Generation & \\
\midrule
\rowcolor{pegray!9}
\multicolumn{7}{c}{\textit{T2I backbone: FLUX.2-dev}} \\
Original Request & 50.84 & 53.21 & 51.97 & 43.84 & 47.59 & 50.52 \\
+ Qwen-Image-PE-9B & 51.77 & 57.38 & 54.21 & 45.51 & 51.71 & 53.05 \\
+ Qwen-Image-PE-397B & \textbf{52.17} & \textbf{58.16} & \textbf{55.85} & \textbf{47.00} & \textbf{55.03} & \textbf{54.38} \\
\rowcolor{pegray!9}
\multicolumn{7}{c}{\textit{T2I backbone: Krea 2 Turbo}} \\
Original Request & 51.71 & 50.31 & 50.58 & 46.47 & 46.27 & 49.76 \\
+ Qwen-Image-PE-9B & 53.44 & 58.54 & 56.50 & 47.55 & 55.40 & 55.06 \\
+ Qwen-Image-PE-397B & \textbf{53.99} & \textbf{59.96} & \textbf{57.96} & \textbf{48.61} & \textbf{57.08} & \textbf{56.28} \\
\rowcolor{pegray!9}
\multicolumn{7}{c}{\textit{T2I backbone: HunyuanImage-2.1}} \\
Original Request & 45.19 & 46.46 & 46.11 & 41.51 & 39.33 & 44.48 \\
+ HunyuanImage-2.1-PE-32B & 47.08 & 50.69 & 51.19 & 42.82 & 47.49 & 48.59 \\
+ Qwen-Image-PE-9B & \textbf{49.79} & 54.31 & 52.60 & 43.95 & 50.22 & 51.04 \\
+ Qwen-Image-PE-397B & 49.48 & \textbf{54.47} & \textbf{54.28} & \textbf{44.85} & \textbf{51.63} & \textbf{51.66} \\
\rowcolor{pegray!9}
\multicolumn{7}{c}{\textit{T2I backbone: FLUX 3 Image Klein}} \\
Original Request & 51.74 & 53.27 & 51.61 & 45.72 & 49.30 & 51.07 \\
+ FLUX 3 Image Klein PE & 52.84 & 58.25 & 58.92 & 49.61 & 60.56 & 56.36 \\
+ Qwen-Image-PE-9B & \textbf{55.43} & 60.89 & 59.21 & 51.09 & 59.62 & 57.81 \\
+ Qwen-Image-PE-397B & 55.27 & \textbf{62.59} & \textbf{60.78} & \textbf{51.63} & \textbf{63.24} & \textbf{59.10} \\
\bottomrule
\end{tabular}%
}
\caption{\textbf{Generalization behavior of PE.} Both PE variants outperform original request on four T2I backbones, surpassing native enhancers on HunyuanImage-2.1 and FLUX 3 Image Klein.}
\label{tab:backbone_ablation}
\label{tab:backbone_generalization}
\end{table}

\Needspace{12\baselineskip}
\section{Generalization Behavior of Qwen-Image-PE}
\label{sec:generalization}

To assess transfer beyond the training backbone, Qwen-Image-2.1, we apply \mbox{Qwen-Image-PE-9B} and \mbox{Qwen-Image-PE-397B} to four frozen T2I backbones: FLUX.2-dev, Krea 2 Turbo, HunyuanImage-2.1, and FLUX 3 Image Klein. We evaluate on Qwen-Image-Bench against original requests and the available native prompt enhancers.

\paragraph{Consistent improvements across backbones.}
As shown in \Cref{tab:backbone_generalization}, both variants outperform original requests on all four backbones, with improvements across all five benchmark dimensions. Overall improves by 2.53--6.74 points with \mbox{Qwen-Image-PE-9B} and by 3.86--8.03 points with \mbox{Qwen-Image-PE-397B}. These consistent improvements show that the learned prompt enhancement capability generalizes beyond the generator used during training.

\paragraph{Transfer beyond native prompt enhancement.}
Interestingly, both variants surpass the native prompt enhancers in Overall on HunyuanImage-2.1 and FLUX 3 Image Klein. \mbox{Qwen-Image-PE-397B} exceeds HunyuanImage-2.1-PE-32B and FLUX 3 Image Klein PE by 3.07 and 2.74 points, respectively. Achieving these gains without access to the target generators' training captions suggests that Qwen-Image-PE learns visual design principles that transfer across generators.

\FloatBarrier
\section{Related Work}
\label{sec:related}

\paragraph{Text-to-Image Generation.}
Text-to-image (T2I) generation aims to synthesize images that correspond to textual descriptions. Earlier work advanced image fidelity and text--image alignment, enabling models to generate visually plausible subjects and scenes from text~\citep{rombach2022high,podell2023sdxl}. More recent systems support increasingly precise control over visual presentation, allowing detailed prompts to guide lighting, composition, layout, and fine-grained detail~\citep{wu2025qwen,Zhao2026QwenImage20TR,bytedance_seedream5_lite_2025,blackforestlabs2025flux2,openai2026images2,google2026nanobanana2}. Although these models support detailed visual control, user requests often leave design choices unspecified. Our work studies how prompt enhancement can specify and coordinate these choices through detailed prompts to form coherent visual designs.

\paragraph{Prompt Enhancement for Text-to-Image Generation.}
Prompt enhancement (PE) transforms user requests into detailed prompts while preserving user intent. A central objective is semantic alignment, which involves clarifying user intent and expanding generation conditions while preserving explicit content and spatial constraints~\citep{booguimage,blackforestlabs2026flux3image,structuredprompt}. HunyuanImage-2.1-PE further employs chain-of-thought reasoning to produce detailed prompts that faithfully express user requirements~\citep{promptenhancer}. Beyond semantic correctness, existing approaches optimize visual appeal through aesthetic and preference rewards~\citep{promptist,beautifulprompt,prism2026,poet}. ERNIE-Image-PE also enriches requests with structured descriptions of layout, composition, textual content, and scene structure~\citep{ernieimage}.

Our work optimizes fine-grained visual design dimensions, including lighting, composition, layout, and detail. Supervised fine-tuning with Caption-Based Reverse Distillation trains the enhancer to imitate design choices captured in dense captions of well-designed images, while reinforcement learning with Groupwise Design Reward refines these choices through groupwise visual comparison. Concurrent work WanPE~\citep{zhu2026wanpe} also constructs supervision from captions for text-to-video prompt enhancement. It targets coherent cinematic planning across shots and over time, whereas our work focuses on fine-grained visual design within individual images. In addition, we study the scaling properties of PE and its generalization behavior across T2I backbones.

\section{Conclusion}

We present Qwen-Image-PE, which learns fine-grained visual design through imitation and comparative refinement. Supervised fine-tuning via Caption-Based Reverse Distillation provides concrete design supervision, while reinforcement learning with Groupwise Design Reward refines design choices through groupwise visual comparison. Experiments show improvements in overall generation performance, visual design, and human preference. Our scaling experiments suggest that sufficient information density is a prerequisite for effective enhancement. Both PE variants also improve Overall over original requests on four additional T2I backbones without access to their training captions, supporting the transfer of learned visual design principles. We hope this work encourages future research to make visual design a central objective of prompt enhancement.

\bibliography{colm2024_conference}
\bibliographystyle{colm2024_conference}
\end{document}